\documentclass[a4paper,10pt]{article}

\usepackage{amssymb,amsmath,array,bm}

\usepackage{graphicx}
\usepackage{textcomp}
\usepackage{xcolor}

\usepackage{booktabs}

\usepackage{algorithm}
\usepackage{algpseudocode}

\usepackage{tikz}
\usepackage{tikz-qtree}
\usetikzlibrary{arrows}
\tikzstyle{edge}=[->, >=stealth', shorten <=2pt, shorten >=2pt, auto, semithick]
\usepackage{tango-colors}

\usepackage{pgfplots}
\pgfplotsset{compat=1.14}
\usepgfplotslibrary{groupplots}

\usepackage{hyperref}
\usepackage[utf8]{inputenc}
\usepackage[T1]{fontenc}
\usepackage[english]{babel}

\usepackage{tango-colors}

\usepackage[a4paper, left=2.5cm, right=2.5cm, top=3cm, bottom=3cm]{geometry}

\usepackage{authblk}
\usepackage{natbib}
\usepackage{doi}

\usepackage[utf8]{inputenc}
\usepackage[T1]{fontenc}
\usepackage[english]{babel}
\usepackage{csquotes}

\newcommand{\R}{\mathbb{R}}
\newcommand{\N}{\mathbb{N}}
\newcommand{\enc}{f}
\newcommand{\dec}{g}
\newcommand{\Enc}{\phi}
\newcommand{\Dec}{\psi}
\newcommand{\out}{h}
\newcommand{\tre}[1]{\hat{#1}}

\renewcommand{\cite}{\citep}
\newenvironment{IEEEkeywords}{\paragraph{Keywords:}}{}

\author{Benjamin Paaßen}
\author{Irena Koprinska}
\author{Kalina Yacef}
\affil{School of Computer Science, The University of Sydney\thanks{Funding by the German Research Foundation (DFG) under grant number PA 3460/1-1 is gratefully acknowledged. Online supplement with source code at \url{https://gitlab.com/bpaassen/tree_echo_state_autoencoders}.}}

\title{Tree Echo State Autoencoders with Grammars}

\date{Preprint of the IJCNN 2020 paper \citet{Paassen2020IJCNN} as provided by the authors.%
\footnote{© 2020 IEEE. Personal use of this material is permitted.  Permission from IEEE must be obtained for all other uses, in any current or future media, including reprinting/republishing this material for advertising or promotional purposes, creating new collective works, for resale or redistribution to servers or lists, or reuse of any copyrighted component of this work in other works.}}

\begin{document}

\maketitle

\pagestyle{myheadings}
\markright{Preprint of \citet{Paassen2020IJCNN} provided by the authors.}

\begin{abstract}
Tree data occurs in many forms, such as computer programs, chemical molecules,
or natural language. Unfortunately, the non-vectorial and discrete nature of
trees makes it challenging to construct functions with tree-formed output, complicating
tasks such as optimization or time series prediction. Autoencoders address this challenge
by mapping trees to a vectorial latent space, where tasks
are easier to solve, and then mapping the solution back to a tree structure.
However, existing autoencoding approaches for tree data fail to take
the specific grammatical structure of tree domains into account and rely
on deep learning, thus requiring large training datasets
and long training times. In this paper, we propose tree echo state autoencoders
(TES-AE), which are guided by a tree grammar and can be trained within seconds by virtue
of reservoir computing. In our evaluation on three datasets, we demonstrate
that our proposed approach is not only much faster than a state-of-the-art
deep learning autoencoding approach (D-VAE) but also has less autoencoding error
if little data and time is given.
\end{abstract}


\section{Introduction}

Trees constitute an important data structure in a wide range of fields,
describing diverse data such as computer programs \cite{Aho2006}, chemical
molecules \cite{Weininger1988}, or natural language \cite{Knight2005}. 
In recent years, machine learning on these kinds of data has made considerable process,
especially for classification and regression tasks \cite{Gallicchio2013,Kipf2017,Paassen2018ICML}.
In these cases, a machine learning model maps from trees to a scalar or vectorial output (\emph{encoding}).
The converse direction, mapping a vector back to a tree (\emph{decoding}), however, is
less well investigated, although such decoders would be highly useful for tasks such as
generative models for trees, the optimization of tree structures, or time series prediction
on trees \cite{Paassen2019ESANN}. In particular, a decoder for trees could help to optimize
molecular structures \cite{Kusner2017}, or to provide hints to students in intelligent
tutoring systems \cite{Paassen2018JEDM}.

Prior work on decoders for structured data can be roughly partitioned into two groups.
First, decoders for full or acyclic graphs \cite{Liu2018,You2018,Bacciu2019,Zhang2019}, which
use deep recurrent neural networks to generate a graph one node or edge at a time until
a full graph is completed. The drawback of these approaches is that they fail to take the
specific structure of trees into account and thus may generate structures that are not trees.
Furthermore, they do not take grammatical knowledge about the domain into account, which would
be available for all aforementioned examples \cite{Aho2006,Weininger1988,Knight2005}, and could
thus be a useful prior.

The second group are decoders that take grammar information into account
\cite{Kusner2017,Dai2018}, but are at present limited to string data instead of trees.
Furthermore, both groups rely on deep neural networks for training which require large datasets
and long training times.

Our key contribution in this paper are tree echo state autoencoders (TES-AE),
a novel autoencoder architecture specifically dedicated to tree data, which uses
grammatical knowledge and can be trained within seconds using a standard support vector machine solver \cite{Cortes1995,Scikit2011}. Our approach is
based on tree echo state networks \cite{Gallicchio2013} for encoding and analogous networks
for decoding, where we keep all neural network parameters fixed except for the final decoding
layer. In our proposed model, this final layer decides which grammar rule to apply
in each step of the decoding process.
In our experiments on three datasets we show that our autoencoding approach can outperform
deep variational autoencoders for acyclic graphs (D-VAE) \cite{Zhang2019} in terms of training
time and autoencoding error, if little data and little training time is available.
Further, we show that TES-AEs outperform sequential echo state networks for this application
and that the TES-AE coding space is suitable for tree optimization, achieving similar results
as \cite{Kusner2017}.

In the following, we cover related work in more detail and recap background knowledge regarding
regular tree grammars, before we describe our proposed architecture in depth, explain our
experiments and results, and conclude with a summary of our findings.

\section{Related Work}

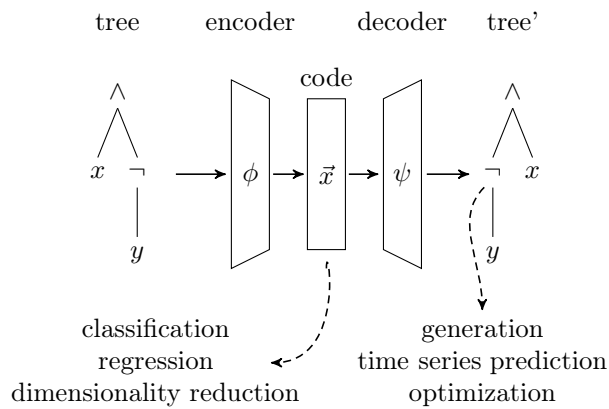
\begin{figure}
\begin{center}
\begin{tikzpicture}

\begin{scope}[shift={(-1.5,+1)}]
\Tree [.$\wedge$ $x$ [.$\neg$ $y$ ] ]
\end{scope}
\node at (-1.5,2) {$\strut$tree};

\draw (0,1.25) -- (0.5,1) -- (0.5,-1) -- (0,-1.25) -- cycle;
\node at (0.25,2) {$\strut$encoder};
\node (enc) at (0.25,0) {$\Enc$};

\draw[edge] (-0.8,0) -- (0,0);

\draw (1,1) rectangle (1.5,-1);
\node at (1.25,1.25) {$\strut$code};
\node (code) at (1.25,0) {$\vec x$};

\path[edge] (enc) edge (code);

\node[align=center] (app_enc) at (-1,-2.5) {classification\\regression\\dimensionality reduction};

\path[edge,densely dashed, shorten <= 1cm, shorten >=-0.5cm] (code) edge[out=-90, in=0, looseness=0.6] (app_enc);

\draw (2,1) -- (2.5,1.25) -- (2.5,-1.25) -- (2,-1) -- cycle;
\node at (2.25,2) {$\strut$decoder};
\node (dec) at (2.25,0) {$\Dec$};

\path[edge] (code) edge (dec);

\begin{scope}[shift={(+3.7,+1)}]
\Tree [.$\wedge$ [.$\neg$ $y$ ] $x$ ]
\end{scope}
\node at (3.7,2) {$\strut$tree'};

\draw[edge] (2.5,0) -- (3.2,0);

\node[align=center] (app_dec) at (3.3,-2.5) {generation\\time series prediction\\optimization};

\node (tree2) at (3.5,0) {};
\path[edge, densely dashed, shorten >=0pt] (tree2) edge[out=225,in=90] (app_dec);

\end{tikzpicture}
\end{center}
\caption{An illustration of a tree encoder/decoder architecture and the
applications for the code (classification, regression, dimensionality reduction)
and the decoded tree (generation, optimization, time series prediction).}
\label{fig:tree_auto_encoder}
\end{figure}

\subsection{Tree Encoding}

Most prior work on machine learning for trees can be grouped into neural network approaches
(e.g.\ \cite{Sperduti1997,Hammer2004,Gallicchio2013}) and tree kernel approaches (e.g.\ \cite{Collins2002,Aiolli2015}).
In both cases, a tree $\tre x$ is first mapped to a vectorial representation $\Enc(\tre x) = \vec x$, which is then used to complete
a machine learning task, such as classification \cite{Sperduti1997}, regression \cite{Gallicchio2013},
or dimensionality reduction \cite{Hammer2004}. We call the mapping $\Enc$ an \emph{encoder} for trees
and we call $\vec x$ the \emph{code} of $\tre x$ (refer to Figure~\ref{fig:tree_auto_encoder}, left).
In more detail, recursive neural networks \cite{Sperduti1997,Hammer2004,Gallicchio2013} encode trees
by defining a function $\enc$ which maps a node label and a (perhaps ordered) set of child encodings
to an encoding for the parent node. The overall encoding $\Enc$ is then computed via recursion.
For example, the tree $\tre x = \wedge(x, \neg(y))$ from Figure~\ref{fig:tree_auto_encoder}
would be recursively encoded as $\Enc(\tre x) = \enc(\wedge, \{ \Enc(x), \Enc(\neg(y))\})$,
where $\Enc(x) = \enc(x, \emptyset)$ and $\Enc(\neg(y)) = \enc(\neg, \{\Enc(y)\}) = \enc(\neg, \{ \enc(y, \emptyset) \})$.
We follow this recursive encoding scheme in our work but adapt it slightly to be better aligned with a grammar.

\subsection{Tree Decoding}

While an encoder is sufficient to perform machine learning tasks with vectorial output,
many interesting tasks require a \emph{decoder} $\Dec$ as well, i.e.\ a mapping from the vector
space back to the space of trees (refer to Figure~\ref{fig:tree_auto_encoder}, right).
For example, we can address time series prediction by encoding a tree $\tre x$ as a vector $\Enc(\tre x)$,
predicting the next state of the vector $\Enc(\tre x) + \vec \delta$, and then decoding back to the next state of the tree
$\Dec(\Enc(\tre x) + \vec \delta)$
\cite{Paassen2018JEDM}; we can construct new trees by sampling a vector $\vec x$ in the latent space and then
mapping back to a tree structure $\Dec(\vec x)$ \cite{Bacciu2019}; and we can optimize trees by
varying the representation in the latent space $\vec x$ such that some objective function $\ell(\Dec(\vec x))$
on the decoded tree $\Dec(\vec x)$ is optimized \cite{Kusner2017}.

Training a decoder for trees is considerably harder compared to an encoder because
the dimensionality of the vector space (and hence the number of neurons in the model)
needs to scale exponentially with the tree depth to distinguish all possible trees in a domain \cite{Hammer2002}.
Accordingly, only few scholars to date have attempted to tackle the problem of tree decoding \cite{Paassen2019ESANN}.
Most who did are concerned with the more general problem of graph decoding by
generating a graph one node/edge at a time via a deep recurrent neural network \cite{Liu2018,You2018,Bacciu2019,Zhang2019}.
In more detail, these approaches treat a graph as a sequence of node and edge insertions and
attempt to reproduce this sequence with a recurrent neural network.
The most applicable of these works to our setting are variational autoencoders for
directed acyclic graphs (D-VAE) \cite{Zhang2019} because trees are a subclass of
acyclic graphs and thus the architectural bias towards acyclicity should help
D-VAEs in reconstructing trees.

Note that our proposed model is similar to these approaches in that we also equate
a tree with a sequence of actions, namely a sequence of production rules in a regular
tree grammar. However, we do not apply a recurrent neural network but follow the
recursive structure of the tree. Further, by considering grammar rules instead of
general node and edge insertions, our output trees are guaranteed to be syntactically
correct whereas existing graph decoders may violate syntactic rules or produce data
that is not tree-formed at all.

With respect to the reliance on grammars, our approach resembles the work of \cite{Kusner2017}
who also suggested to guide a decoder by a grammar. Also like \cite{Kusner2017}, we train
our networks to achieve autoencoding, i.e.\ we wish to train a $\Dec$ that acts as an inverse
of an encoder $\Enc$ on the training data. However, we consider tree data instead of string
data and use recursive networks instead of (time-)convolutional networks.

\subsection{Echo State Networks}

A final and crucial difference to all previous work lies in our choice of training scheme.
While all aforementioned approaches use gradient descent across the entire network, we base
our approach on the reservoir computing literature (e.g.\ \cite{Jaeger2004,Rodan2012}).
More precisely, we use a slightly varied version of the tree echo state network \cite{Gallicchio2013}
as encoder and decoder, where all internal parameters are initialized randomly, then
pre-processed to ensure eventual forgetting of inputs \cite{Jaeger2004}, but kept fixed
afterwards. We only train the final layer that decides which grammar rule to take in each
step of the decoding. Because of this, our training problem becomes convex and easy to solve.
In particular, we can use a straightforward support vector machine solver \cite{Cortes1995,Scikit2011} to
train the output layer. Our main contribution to the reservoir computing literature is
that we propose not only an encoder, but a decoder model for trees.

\subsection{Regular tree grammars}

Our approach strongly relies on regular tree grammars \cite{Brainerd1969,Comon2008},
such that we now take some time to describe them in more detail, albeit in a
simplified notation to ease understanding.

First, we define a \emph{tree} $\tre x$ over some finite alphabet $\Sigma$ as an expression
$x(\tre y_1, \ldots, \tre y_k)$, where $x \in \Sigma$ and where $\tre y_1, \ldots, \tre y_k$
are also trees over $\Sigma$, which we call the \emph{children} of $\tre x$. Note that $k$ may
be zero, in which case we call the tree a \emph{leaf}.
For example, for $\Sigma = \{ \wedge, \vee, \neg, x, y \}$, $x()$,
$\vee(x(), y())$, $\neg(x())$, and $\wedge()$ are all trees over $\Sigma$, where $x()$
and $\wedge()$ are leaves. Per convention, we omit the empty brackets for leaves.

Note that our definition of trees is very liberal and includes many instances that
may be nonsensical according to the rules of the domain. To restrict the space of
possible trees to a more sensible subset, we use \emph{regular tree grammars}.
We define a regular tree grammar as a $4$-tuple $\mathcal{G} = (\Phi, \Sigma, R, S)$,
where $\Phi$ is a finite set of nonterminal symbols, $\Sigma$ is a finite set of
terminal symbols, $S \in \Phi$ is a special nonterminal symbol which we call the
\emph{starting} symbol, and $R$ is a finite set of production rules of the form
$A \to x(B_1, \ldots, B_k)$ where $A, B_1, \ldots, B_k \in \Phi$ and $x \in \Sigma$.

We say that a tree $\tre y$ over $\Phi \cup \Sigma$ can be \emph{derived in one step}
via grammar $\mathcal{G}$ from another tree $\tre x$ over $\Phi \cup \Sigma$, if there
exists a production rule $A \to x(B_1, \ldots, B_k)$ and a leaf $A$ in $\tre x$,
such that replacing $A$ with $x(B_1, \ldots, B_k)$ yields $\tre y$. Generalizing this
definition, we say that a tree $\tre y$ can be derived in $T$ steps via grammar $\mathcal{G}$
from another tree $\tre x$, if there exists a sequence of trees $\tre z_0 \to \ldots \to
\tre z_T$ such that $\tre z_0 = \tre x$, $\tre z_T = \tre y$, and $\tre z_t$ can
be derived in one step via grammar $\mathcal{G}$ from $\tre z_{t-1}$ for all
$t > 0$. Finally, we define the \emph{tree language} $\mathcal{L}(\mathcal{G})$
as the set of all trees $\tre x$ over $\Sigma$
which can be derived in $T$ steps from the starting symbol $S$ for any $T \in \N$.
As an example, consider the regular tree grammar in Figure~\ref{fig:encoding}, left.
The tree $\wedge(x, \neg(y))$ can be derived in $4$ steps from $S$ via
the sequence $S \to \wedge(S, S) \to \wedge(x, S) \to \wedge(x, \neg(S)) \to \wedge(x, \neg(y))$.

An important property of regular tree grammars is that they can be parsed efficiently
using tree automata \cite{Brainerd1969,Comon2008}. This is especially easy to see for a subclass
of regular tree grammars, which we call \emph{deterministic}.
We define a regular tree grammar as deterministic if no two production
rules have the same right-hand-side. For these grammars, we can parse a tree $\tre x
= x(\tre y_1, \ldots, y_k)$ via the following recursive function: First, we parse all
children of $\tre x$. This will return a nonterminal symbol $B_i$ for every child
$\tre y_i$ and a sequence of rules deriving $\tre y_i$ from $B_i$. After that, we
simply have to check whether a rule of the form $A \to x(B_1, \ldots, B_k)$ exists
in our grammar. If so, we return the nonterminal symbol $A$ and the concatenation of
this rule and all rule sequences for the children. If not, the parse ends because the
tree is not part of the tree language. We utilize this scheme later for encoding
in Algorithm~\ref{alg:encoding}.

\section{Method}
\label{sec:methods}

Our aim in this paper is to construct an autoencoder for trees that exploits grammatical
knowledge for the tree domain. More precisely, for a given regular tree grammar $\mathcal{G}$
we would like to obtain an encoder $\Enc : \mathcal{L}(\mathcal{G}) \to \R^n$ for some $n \in \N$
and a decoder $\Dec : \R^n \to \mathcal{L}(\mathcal{G})$, such that for as many trees $\tre x \in \mathcal{L}(\mathcal{G})$
as possible, $\tre x$ is close to $\Dec(\Enc(\tre x))$.
To achieve this goal, we introduce two approaches. We start with a sequence-to-sequence
learning approach following the architecture of \cite{Sutskever2014} and then continue
with an approach based on tree echo state networks \cite{Gallicchio2013},
which we describe in terms of encoding, decoding, and training.

\subsection{Sequence-to-sequence learning}
\label{sec:seq2seq}

Sequence-to-sequence learning is a neural network architecture introduced by \cite{Sutskever2014},
which translates an input sequence to an output sequence, potentially of different length.
The architecture features two recurrent neural networks, an encoding network
$\enc : \R^l \times \R^n \to \R^n$, a decoding network $\dec : \R^l \times \R^n \to \R^n$,
and an output function $\out : \R^n \to \R^l$ for some input dimensionality $l$ and
encoding dimensionality $n$.
The encoding network translates the input time series $\vec y_1, \ldots, \vec y_T \in \R^l$ into an encoding
vector $\vec x_T$ by means of the equation $\vec x_t = \enc(\vec y_t, \vec x_{t-1})$ where
$\vec x_0 = \vec 0$, i.e.\ a vector of $n$ zeros. This encoding is then used to generate the output time series
$\vec z_1, \ldots, \vec z_{T'}$ as follows. We first set the initial decoding state as $\tilde x_1 = \vec x_T$
and then generate the first output as $\vec z_1 = h(\tilde x_1)$. All remaining decoding states are
generated via $\tilde x_t = \dec(\vec z_{t-1}, \tilde x_{t-1})$ and all remaining outputs via 
$\vec z_t = h(\tilde x_t)$ until $\vec z_t$ is a special end-of-sequence token, whereupon the
process stops.

To apply the sequence-to-sequence learning framework to tree data, we first translate an input tree
$\tre x$ into a sequence of production rules, which we represent via one-hot codes, then encode this sequence
via the encoder network, decode it to a sequence of one-hot codes, translate these back to production
rules, and finally produce the decoded tree using these rules.

\begin{figure*}
\begin{center}
\begin{tikzpicture}

\node (y1) at (0,2) {$S \to \wedge(S, S)$};
\node (y2) at (2,2) {$S \to x$};
\node (y3) at (4,2) {$S \to \neg(S)$};
\node (y4) at (6,2) {$S \to y$};

\node (y1onehot) at (0,1) {\footnotesize $(1, 0, 0, 0, 0, 0)$};
\node (y2onehot) at (2,1) {\footnotesize $(0, 0, 0, 1, 0, 0)$};
\node (y3onehot) at (4,1) {\footnotesize $(0, 0, 1, 0, 0, 0)$};
\node (y4onehot) at (6,1) {\footnotesize $(0, 0, 0, 0, 1, 0)$};

\node (x0) at (-2,0) {$\vec x_0 = \vec 0$};
\node (x1) at (0,0)  {$\vec x_1$};
\node (x2) at (2,0)  {$\vec x_2$};
\node (x3) at (4,0)  {$\vec x_3$};
\node (x4) at (6,0)  {$\vec x_4 = \tilde x_1$};

\node (xtilde2) at (8,0)  {$\tilde x_2$};
\node (xtilde3) at (10,0) {$\tilde x_3$};
\node (xtilde4) at (12,0) {$\tilde x_4$};
\node (xtilde5) at (14,0) {$\tilde x_5$};

\node (z1onehot) at (6,-1)  {\footnotesize $(1, 0, 0, 0, 0, 0)$};
\node (z2onehot) at (8,-1)  {\footnotesize $(0, 0, 0, 1, 0, 0)$};
\node (z3onehot) at (10,-1) {\footnotesize $(0, 0, 1, 0, 0, 0)$};
\node (z4onehot) at (12,-1) {\footnotesize $(0, 0, 0, 0, 1, 0)$};
\node (z5onehot) at (14,-1) {\footnotesize $(0, 0, 0, 0, 0, 1)$};

\node (z1) at (6,-2) {$S \to \wedge(S, S)$};
\node (z2) at (8,-2) {$S \to x$};
\node (z3) at (10,-2) {$S \to \neg(S)$};
\node (z4) at (12,-2) {$S \to y$};
\node (z5) at (14,-2) {EOS};

\path[edge]
% rule seq to one hot coding
(y1) edge (y1onehot)
(y2) edge (y2onehot)
(y3) edge (y3onehot)
(y4) edge (y4onehot)
% one hot coding to state
(y1onehot) edge node[left, pos=0.9, outer sep=0.3cm] {$\enc$} (x1)
(y2onehot) edge node[left, pos=0.9, outer sep=0.3cm] {$\enc$} (x2)
(y3onehot) edge node[left, pos=0.9, outer sep=0.3cm] {$\enc$} (x3)
(y4onehot) edge node[left, pos=0.9, outer sep=0.3cm] {$\enc$} (x4)
% encoding state-to-state
(x0) edge (x1)
(x1) edge (x2)
(x2) edge (x3)
(x3) edge (x4)
% decoding state-to-state
(x4) edge (xtilde2)
(xtilde2) edge (xtilde3)
(xtilde3) edge (xtilde4)
(xtilde4) edge (xtilde5)
% state to output
(x4) edge node[left] {$\out$} (z1onehot)
(xtilde2) edge node[left] {$\out$} (z2onehot)
(xtilde3) edge node[left] {$\out$} (z3onehot)
(xtilde4) edge node[left] {$\out$} (z4onehot)
(xtilde5) edge node[left] {$\out$} (z5onehot)
% output to state
(z1onehot) edge node[left, pos=0.9, outer sep=0.3cm] {$\dec$} (xtilde2)
(z2onehot) edge node[left, pos=0.9, outer sep=0.3cm] {$\dec$} (xtilde3)
(z3onehot) edge node[left, pos=0.9, outer sep=0.3cm] {$\dec$} (xtilde4)
(z4onehot) edge node[left, pos=0.9, outer sep=0.3cm] {$\dec$} (xtilde5)
% one hot coding to rule seq
(z1onehot) edge (z1)
(z2onehot) edge (z2)
(z3onehot) edge (z3)
(z4onehot) edge (z4)
(z5onehot) edge (z5);

\end{tikzpicture}
\end{center}
\caption{An illustration of the sequence-to-sequence autoencoding architecture for the example tree
$\wedge(x, \neg(y))$ and the regular tree grammar from Figure~\ref{fig:encoding}. Top: The rule sequence
generating the tree; second row: the translation of the rule sequence into one-hot-coding; third row: the
sequence of encoding and decoding states; last row: the output series of one-hot codings.}
\label{fig:seq2seq}
\end{figure*}
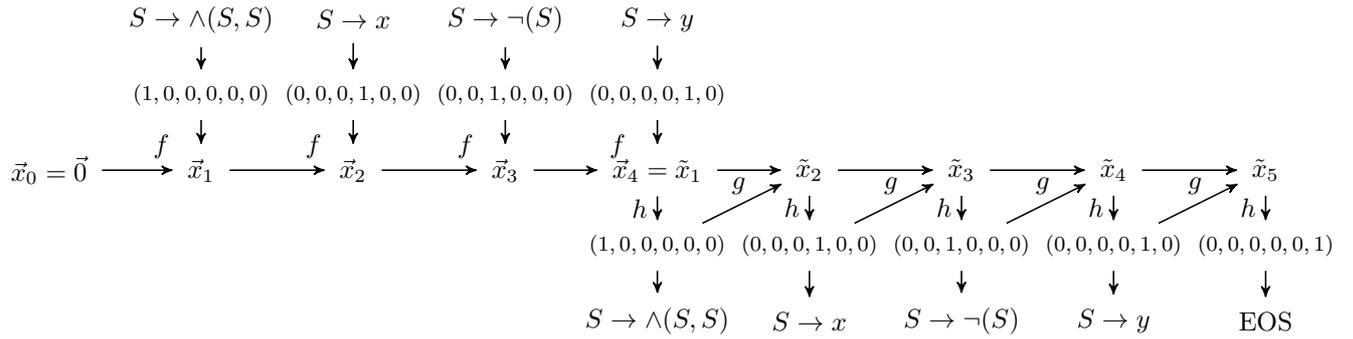

Figure~\ref{fig:seq2seq} illustrates the approach for the example tree
$\wedge(x, \neg(y))$ and the example grammar from Figure~\ref{fig:encoding}.
Recall that our example tree can be derived from the starting symbol $S$ via the sequence
$S \to^{\bm{1}} \wedge(S, S) \to^{\bm{4}} \wedge(x, S) \to^{\bm{3}} \wedge(x, \neg(S))
\to^{\bm{5}} \wedge(x, \neg(y))$, where we indexed each arrow by its corresponding production rule
according to the numbering from Figure~\ref{fig:encoding}. Accordingly, the tree is equivalent to
the production rule sequence $(\bm{1}, \bm{4}, \bm{3}, \bm{5})$, which we represent by one-hot
codes in the second row of Figure~\ref{fig:seq2seq}. We then apply the encoding network $\enc$
four times to achieve an overall encoding $\vec x_4$ of our input sequence, which we then plug
in our decoder as initial state $\tilde x_1$. From this initial state, our output function
$\out$ predicts the first element $\vec y_1$ of our output rule sequence, which is then fed back into the
decoding network $\dec$ to generate the second state $\tilde x_2$, and so on until $\out$ predicts the
special end-of-sequence token $(0, 0, 0, 0, 0, 1)$.

In our case, we implement both $\enc$ and $\dec$ as recurrent neural networks with the equations
$\vec x_t = \enc(\vec y_t, \vec x_{t-1}) = \tanh\big( \bm{U} \cdot \vec y_t + \bm{W} \cdot \vec x_{t-1}\big)$
and $\tilde x_t = \dec(\vec z_{t-1}, \tilde x_{t-1}) = \tanh\big(\bm{U} \cdot \vec z_{t-1} + \bm{W} \cdot \tilde x_{t-1}\big)$,
and the output function as a linear function $\vec z_t = \out(\tilde x_t) = \bm{V} \cdot \tilde x_t$.
Note that the matrices $\bm{U}$, $\bm{W}$, and $\bm{V}$ are parameters of our model. Following the
reservoir computing paradigm \cite{Jaeger2004}, we do not train the matrices $\bm{U}$ or $\bm{W}$ but
initialize them as cycle reservoir with jumps \cite{Rodan2012} and then keep them fixed.
Note that we use the same matrices $\bm{U}$ and $\bm{W}$ for $\enc$ and $\dec$.
Next, we generate for each tree in the training data the decoding state sequence
$\tilde x_1, \ldots, \tilde x_{T+1}$ via teacher forcing, i.e.\
$\tilde x_t = \tanh\big(\bm{U} \cdot \vec y_{t-1} + \bm{W} \cdot \tilde x_{t-1}\big)$,
using $\vec y_{t-1}$ as input argument instead of $\vec z_{t-1}$.
Finally, we train the matrix $\bm{V}$ via linear regression on the training data
$\big\{ (\tilde x_t, \vec y_t) | t \in \{1, \ldots, T\} \big\}$.

While this approach is already functional in principle, we expect it to fail for reasonably large
input trees. This is because our network needs to remember rule applications a long time ago to correctly
predict the next production rule. Echo state networks, however, focus on intense short-term
memory instead of long-term memory \cite{Jaeger2004,Farkas2016}. Accordingly,
we now attempt to reduce the number of time steps between encoding and decoding by working along
the tree structure instead of flattening it to a sequence beforehand.

\subsection{Tree Encoding}

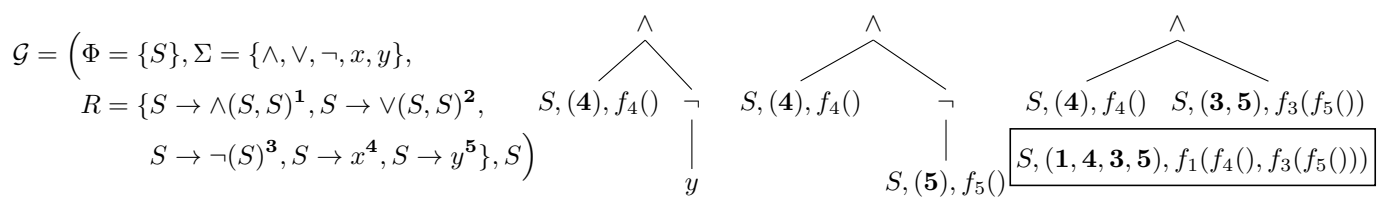
\begin{figure*}
\begin{center}
\begin{tikzpicture}
\node[align=left] at (0.1,0) {$\begin{aligned}
\mathcal{G} = \Big( \Phi = \{ &S \},
\Sigma = \{ \wedge, \vee, \neg, x, y \},\\
R = \{ &S \to \wedge(S, S)^{\bm{1}}, S \to \vee(S, S)^{\bm{2}},\\
&S \to \neg(S)^{\bm{3}}, S \to x^{\bm{4}}, S \to y^{\bm{5}} \}, S\Big)
\end{aligned}$};

\begin{scope}[shift={(+5,0)}]
% \begin{scope}[shift={(0,+1)}]
% \Tree [.$\wedge$ $x$ [.$\neg$ $y$ ] ]
% \end{scope}
\begin{scope}[shift={(0,+1)}]
\Tree [.$\wedge$ {$S, (\bm{4}), \enc_4()$} [.$\neg$ $y$ ] ]
\end{scope}
\begin{scope}[shift={(3,+1)}]
\Tree [.$\wedge$ {$S, (\bm{4}), \enc_4()$} [.$\neg$ {$S, (\bm{5}), \enc_5()$} ] ]
\end{scope}
\begin{scope}[shift={(7,+1)}]
\Tree [.$\wedge$ {$S, (\bm{4}), \enc_4()$} {$S, (\bm{3}, \bm{5}), \enc_3(\enc_5())$} ]
\end{scope}
\begin{scope}[shift={(7.2,-0.8)}]
\Tree [.{$S, (\bm{1}, \bm{4}, \bm{3}, \bm{5}), \enc_1(\enc_4(), \enc_3(\enc_5()))$} ]
\draw[draw=black, semithick] (-2.4,0.5) rectangle (2.4,-0.25);
\end{scope}
\end{scope}

\end{tikzpicture}
\end{center}
\caption{An illustration of the encoding algorithm~\ref{alg:encoding} for the tree
$\wedge(x, \neg(y))$. Left: The tree grammar with enumerated rules (number labels in upper index).
From center to right: Each step of the encoding process with the final result highlighted with a box. During encoding, each node is replaced with a triple of a nonterminal label,
a sequence of grammar rules (here as numbers), and a vectorial encoding (here abstracted via function symbols).}
\label{fig:encoding}
\end{figure*}

To encode a tree, we follow the parsing scheme for (deterministic) regular tree grammars
outlined in the background section. More formally, let $\mathcal{G} = (\Phi, \Sigma, R, S)$
be a regular tree grammar. Then, for each grammar rule $r = \big( A \to x(B_1, \ldots, B_k)\big)
\in R$, we define a function $\enc_r : \R^{n \times k} \to \R^n$, such that we can construct
the encoding $\Enc(\tre x)$ of a tree $\tre x = x(\tre y_1, \ldots, \tre y_k)$ recursively as
\begin{equation}
\Enc(\tre x) = \enc_r\Big( \Enc(\tre y_1), \ldots, \Enc(\tre y_k)\Big)
\end{equation}
The precise algorithm for encoding is outlined in Algorithm~\ref{alg:encoding}.
An example is shown in Figure~\ref{fig:encoding}. In the example, we start with
the entire tree $\wedge(x, \neg(y))$ and pass downward through the tree until we
reach the first leaf, which is $x$. We parse this leaf using the
fourth grammar rule $S \to x$, such that our encoding function returns the
the nonterminal $S$, the rule sequence $(\bm{4})$, and the vector encoding
$\Enc(x) = \enc_4()$.
We perform the same scheme for the leaf $y$, yielding the nonterminal $S$,
the rule sequence $(\bm{5})$, and the encoding $\enc_5()$.
We then proceed with the partially parsed subtree $\neg(S)$, which we can parse using the third rule
$S \to \neg(S)$, yielding the nonterminal $S$, the rule sequence
$(\bm{3}, \bm{5})$, and the encoding $\enc_3(\enc_5())$.
This leaves the tree $\wedge(S, S)$, which we can parse using the first rule,
yielding the nonterminal $S$, the rule sequence $(\bm{1}, \bm{4}, \bm{3}, \bm{5})$, and
the overall encoding $\Enc(\wedge(x, \neg(y))) = \enc_1(\enc_4(), \enc_3(\enc_5()))$.

\begin{algorithm}
\caption{An algorithm to encode and parse trees according to a deterministic regular tree grammar
$\mathcal{G} = (\Phi, \Sigma, R, S)$ and encoding functions $\enc_r$ for each rule $r \in R$.
The algorithm receives a tree as input and returns a nonterminal symbol, a rule sequence
that generates the tree from that nonterminal symbol, and a vectorial encoding.}
\label{alg:encoding}
\begin{algorithmic}
\Function{encode}{a tree $\tre x = x(\tre y_1, \ldots, \tre y_k)$}
\For{$j \in \{1, \ldots, k\}$}
  \State $B_j, (r_{j, 1}, \ldots, r_{j, T_j}), \vec y_j \gets $ \Call{encode}{$\tre y_j$}.
\EndFor
\If{$\exists A \in \Phi : A \to x(B_1, \ldots, B_k) \in R$}
  \State $r \gets \big(A \to x(B_1, \ldots, B_k)\big)$.
  \State \Return $A, ( r, r_{1, 1}, \ldots, r_{k, T_k}), \enc_r(\vec y_1, \ldots, \vec y_k)$.
\Else
  \State Error; $\tre x$ is not in $\mathcal{L}(\mathcal{G})$.
\EndIf
\EndFunction
\end{algorithmic}
\end{algorithm}

We implement each of the functions $\enc_r$ as a single-layer feedforward neural network,
i.e.
\begin{equation}
\enc_r(\vec y_1, \ldots, \vec y_k) = \tanh\Big( \bm{W}^r_1 \cdot \vec y_1 + \ldots + \bm{W}^r_k \cdot \vec y_k + \vec b^r\Big) \label{eq:ffnn}
\end{equation}
where the $n \times n$ matrices $\bm{W}^r_1, \ldots, \bm{W}^r_k$ and the bias vector $\vec b^r \in \R^n$
are parameters of $\enc_r$. Following the reservoir computing paradigm, we do not
train these parameters but keep them fixed \cite{Jaeger2004}. In more detail, we initialize
a $\beta \in (0, 1]$ fraction of the entries for each matrix as standard normally distributed
random numbers, and then enforce a spectral radius of $\rho \in (0, 1)$. We fill the bias vectors
with normally distributed random numbers with zero mean and standard deviation $\rho$.
Note that the coding dimensionality $n$, as well as the sparsity $\beta$ and the spectral radius
$\rho$ are hyper-parameters of our approach.

We remark in passing that the reservoir computing paradigm would suggest that each of the reservoir matrices
$\bm{W}^r_j$ is universal \cite{Jaeger2004,Rodan2012}. Accordingly, one could assume that it suffices to initialize
\emph{one} reservoir matrix and re-use it across the entire model instead of initializing a separate matrix for
each argument of each rule.
However, using the same reservoir for all input arguments collapses Equation~\ref{eq:ffnn} to
$\tanh\big(\bm{W} \cdot (\vec x_1 + \ldots + \vec x_k)\big)$, which is now an order-invariant function
with respect to the input and, as such, strictly less powerful. Still, we will consider this version as a
baseline in our experiments later on.

\subsection{Tree Decoding}

For decoding, we emulate the production process of a regular tree grammar. We begin with the starting
symbol $S$ and the vectorial code $\vec x$ for the tree to be decoded. Then, we let a classifier
$\out_S : \R^n \to R$ decide which of the possible rules $r = S \to x(B_1, \ldots, B_k)$ with $S$
on the left-hand-side we should apply. Next, we decode $\vec x$ into vectorial codes $\vec y_1, \ldots, \vec y_k$
for the children. For this step, we use decoding functions $\dec^r_j : \R^n \to \R^n$ that should extract the
information for the $j$th child from $\vec x$. We then repeat this scheme recursively until all nonterminal
symbols are decoded. We present the decoding scheme more formally in Algorithm~\ref{alg:decoding}.

As an example, consider Figure~\ref{fig:decoding}. We start at the top with the vector code for the entire tree
and the starting nonterminal $S$. The classifier $\out_S$ then selects the first rule $S \to \wedge(S, S)$ (top right) to apply. Based on this selection, we know that we need to use the decoding functions $\dec^{\bm{1}}_1$ and
$\dec^{\bm{1}}_2$ to obtain vectorial codings $\vec y_1$ and $\vec y_2$ for the new children.
We then apply the same scheme to the newly created vector codes and nonterminals until the entire tree is decoded.

\begin{algorithm}
\caption{An algorithm to decode vectors to trees according to a regular tree grammar $\mathcal{G} = (\Phi, \Sigma, R, S)$,
classifiers $\out_A : \R^n \to R$ for each nonterminal $A \in \Phi$, and decoding functions $\dec^r_j$ for each rule
$r \in R$ and each of its arguments $j$. The function receives a vector and a nonterminal symbol as input and returns
a decoded tree.}
\label{alg:decoding}
\begin{algorithmic}
\Function{decode}{a vector $\vec x \in \R^n$, a nonterminal $A \in \Phi$}
\State $r = \big(A \to x(B_1, \ldots, B_k)\big) \gets \out_A(\vec x)$.
\For{$j \in \{1, \ldots, k\}$}
  \State $\vec y_j \gets \dec^r_j(\vec x)$.
  \State $\tre y_j \gets$ \Call{decode}{$\vec y_j$, $B_j$}.
  \State $\vec x \gets \vec x - \vec y_j$.
\EndFor
\State \Return $x(\tre y_1, \ldots, \tre y_k)$.
\EndFunction
\end{algorithmic}
\end{algorithm}

\begin{figure}
\begin{center}
\begin{tikzpicture}

\node (x) at (0,0) {$\vec x, S$};

\node[right] (andrule) at (2.5,0) {$S \to \wedge(S, S)$};
\node (y1) at (-1.5,-1) {$\vec y_1, S$};
\node (y2) at (1.5,-1) {$\vec y_2, S$};

\node[left] (xrule) at (-2.5,-1) {$S \to x$};
\node[right] (notrule) at (2.5,-1) {$S \to \neg(S)$};
\node (y21) at (1.5,-2.2) {$\vec y_{2, 1}, S$};

\node[right] (yrule) at (2.5,-2.2) {$S \to y$};

\path[edge, gray]
(x) edge node[above] {$\out_S$} (andrule)
(y1) edge  node[above] {$\out_S$} (xrule)
(y2) edge  node[above] {$\out_S$} (notrule)
(y21) edge node[above] {$\out_S$} (yrule);
\path[edge]
(x)  edge node[above left, pos=0.8, inner sep=0pt] {$\dec^{\bm{1}}_1$} (y1)
(x)  edge node[above right, pos=0.8, inner sep=0pt] {$\dec^{\bm{1}}_2$} (y2)
(y2) edge node[left] {$\dec^{\bm{3}}_1$} (y21);

\end{tikzpicture}
\end{center}
\caption{An illustration of the decoding algorithm~\ref{alg:decoding} for the tree
$\wedge(x, \neg(y))$ (from top to bottom).}
\label{fig:decoding}
\end{figure}

Just as before, we implement the decoding functions $\dec^r_j : \R^n \to \R^n$ using
single-layer feedforward neural networks, i.e.: $\dec^r_j(\vec x) = \tanh(\bm{W}^r_j \cdot \vec x + \vec b^r_j)$,
where the matrices $\bm{W}^r_j$ and the bias vectors $\vec b^r_j$ are parameters of the model.
We apply the same initialization scheme for the matrices $\bm{W}^r_j$ and the vectors $\vec b^r_j$
as during encoding, and keep the parameters fixed after initialization.

\subsection{Training}

In our model, we only need to train the rule classifiers $\out_A$ for every nonterminal $A$.
For training these classifiers, we need to know the encoding vectors $\vec x$
for every nonterminal during the decoding process. Fortunately, we can compute these vectors for
our training data using teacher forcing. In particular, recall that
Algorithm~\ref{alg:encoding} does not only yield the encoding for the tree, but also a rule sequence
that generates the tree. This sequence contains the desired outputs for all our classifiers.
Furthermore, we can use this sequence to decide which rules to apply during decoding, such that
we can complete the entire decoding process without relying on the classifiers' outputs.
We describe the details of this computation in Algorithm~\ref{alg:training}. Note that this
algorithm executes Algorithm~\ref{alg:encoding} first and then executes a modified version of
Algorithm~\ref{alg:decoding} where the decision of the rule classifiers $\out_A$ is replaced by
the ground truth rule sequence. The training data
sets $\mathcal{D}_A$ can be accumulated across an entire training set of trees and then be used
to train the rule classifiers $\out_A$. In the example from Figure~\ref{fig:decoding},
the training data would be $\mathcal{D}_S = \{ (\vec x, \bm{1}), (\vec y_1, \bm{4}),
(\vec y_2, \bm{3}), (\vec y_{2, 1}, \bm{5})\}$ because we should execute the first rule when
we encounter the encoding $\vec x$, the fourth rule when we encounter $\vec y_1 = \dec^{\bm{1}}_1(\vec x)$,
the third rule when we encounter $\vec y_1 = \dec^{\bm{1}}_2(\vec x - \vec y_1)$, and the fifth
rule when we encounter $\vec y_{2, 1} = \dec^{\bm{3}}_1(\vec y_2)$.

\begin{algorithm}
\caption{An algorithm to generate training data for the rule classifiers $\out_A$ from
a tree $\tre x$ according to a regular tree grammar $\mathcal{G} = (\Phi, \Sigma, R, S)$.
The algorithm receives a tree $\tre x$ as input and returns a set of training data for
each nonterminal symbol $A \in \Phi$.}
\label{alg:training}
\begin{algorithmic}
\Function{train}{a tree $\tre x$}
\State $A, ( r_1, \ldots, r_T), \vec x \gets$ \Call{encode}{$\tre x$}.
\State Initialize a stack $\mathcal{S}$ with $\vec x$ on top.
\State Initialize an empty set $\mathcal{D}_A$ for each $A \in \Phi$.
\For{$t \gets 1, \ldots, T$}
  \State Let $r_t = A \to x(B_1, \ldots, B_k)$.
  \State Pop $\vec x_t$ from the top of $\mathcal{S}$.
  \State Add $(\vec x_t, r_t)$ to $\mathcal{D}_A$.
  \For{$j \gets k, \ldots, 1$}
    \State $\vec y_j \gets \dec^{r_t}_j(\vec x_t)$.
    \State Push $\vec y_j$ onto $\mathcal{S}$.
    \State $\vec x_t \gets \vec x_t - \vec y_j$.
  \EndFor
\EndFor
\State \Return $\{ \mathcal{D}_A | A \in \Phi \}$.
\EndFunction
\end{algorithmic}
\end{algorithm}

As classifiers $\out_A$ for each nonterminal $A \in \Phi$ we employ a standard support vector
machine \cite{Cortes1995}.

\section{Experiments}

In our experimental evaluation we compare four models. First, a variational autoencoder for directed
acyclic graphs (D-VAE) as proposed by \cite{Zhang2019}; second, the sequence-to-sequence autoencoder
from Section~\ref{sec:seq2seq}, which we call echo-state autoencoder (\emph{ES-AE}); third, our tree echo
state auto-enocoder with shared reservoir matrix across all rules (\emph{S-TES-AE}); and fourth, our tree echo state autoencoder
with separate weight matrices for each rule (\emph{TES-AE}). Generally, we expect the D-VAE model to do
better than all our reservoir computing models because it can adjust all weights instead of just the output
weights. However, we expect that training a D-VAE takes much longer. Between our echo state models,
we expect the TES-AE to do better than the S-TES-AE and the S-TES-AE to do better than the ES-AE,
in alignment with our arguments in Section~\ref{sec:methods}.

We evaluate each model on three datasets. First, a dataset of Boolean expressions (\emph{Boolean}),
which we generate by applying random rules of the grammar in Figure~\ref{fig:encoding} until at most
three binary operators (and/or) are present.

Second, a dataset of function expressions (\emph{expressions}) as described by \cite{Kusner2017}.
The grammar for this dataset is $\mathcal{G} = ( \{ S \}, \{ +, *, /, \sin, \exp, x, 1, 2, 3 \},
\{ S \to +(S, S), S \to *(S, S), S \to /(S, S), S \to \sin(S), S \to \exp(S), S \to x, S \to 1,
S \to 2, S \to 3 \}, S)$. We sample expressions by adding one binary operator to one unary operator
to one unary with a binary argument, e.g.\ $3 * x + \sin(x) + \exp(2 / x)$, which is consistent with the training data generated by \cite{Kusner2017}.

Third, a dataset of 51 python programs (\emph{pysort}) implementing the insertion sort algorithm or parts of it.
The dataset can be found in the online supplement\footnote{\url{https://gitlab.com/bpaassen/tree_echo_state_autoencoders}}.
The grammar is the full python language grammar as documented on \url{https://docs.python.org/3/library/ast.html}.
The statistics for all datasets are listed in Table~\ref{tab:data}.

\begin{table}
\caption{Statistics of the three datasets.}
\label{tab:data}
\begin{center}
\begin{tabular}{lccc}
statistic & Boolean & expressions & pysort \\
\cmidrule(lr){1-1} \cmidrule(lr){2-4}
no.\ of trees & $500$ & $500$ & $51$ \\
no.\ of nonterminals $|\Phi|$ & 1 & 1 & 12 \\
no.\ of terminals $|\Sigma|$ & 5 & 9 & 54 \\
no.\ of rules $|R|$ & 5 & 9 & 54 \\
avg.\ tree size & 5.3 & 9.06 & 64.41 
\end{tabular}
\end{center}
\end{table}

For the D-VAE model, we used the authors' reference implementation\footnote{\url{https://github.com/muhanzhang/D-VAE}}.
We implemented all echo state models in python using
\emph{scikit-learn} \cite{Scikit2011} as support vector machine solver.
All implementations are available in the online supplement\footnotemark[1].
We ran all experiments on a consumer-grade laptop with Intel core i7 CPU.

\subsection{Autoencoding}

We first evaluate the models in terms of their capacity for autoencoding.
As measure of performance, we consider the root mean square error (RMSE),
in particular the formula $\sqrt{ \frac{1}{m} \sum_{i=1}^m d(\tre x_i, \Dec[\Enc(\tre x_i)])^2}$,
where $\tre x_i$ are the test trees, $\Enc$ and $\Dec$ are the en- and decoding
functions of the respective model, and $d$ is the tree edit distance \cite{Zhang1989}.

For the D-VAE model we used the same experimental parameters as in the original paper \cite{Zhang2019}
because the long training times made hyperparameter optimization prohibitive. However, we used
less epochs ($50$) and higher learning rate ($10^{-3}$) to further limit training time.
For the echo state models, we fixed the number of neurons to $256$ to achieve a fair
comparison between the models and optimized all other hyperparameters on extra
validation data. In particular, for \emph{Boolean} and \emph{expressions} we
sampled $100$ additional training trees and $100$ additional test trees specifically
for hyperparameter optimization. For \emph{pysort} we randomly removed $5$ training trees and
$5$ test trees from the main dataset for hyperparameter optimization. The optimization itself
was a random search with $50$ trials for \emph{Boolean} and \emph{expressions}
and $20$ trials for \emph{pysort}.
The precise ranges for each hyper-parameter can be found in the online supplement\footnotemark[1].

For the evaluation itself, we performed a cross-validation with $20$ folds on \emph{Boolean} and \emph{expressions}
and $10$ folds on \emph{pysort}. To keep training times manageable, we evaluated
the D-VAE model only once with a $10\%$ test data split.

We report the RMSEs for all models and all datasets in Table~\ref{tab:errors}.
As expected, the S-TES-AE model clearly outperforms the ES-AE model on all data
sets and the TES-AE model outperforms the S-TES-AE model on the first two
datasets. These differences are statistically significant in a Wilcoxon
sign-rank test with $p < 0.05$ after Bonferroni correction.
On the pysort dataset, the performance
of TES-AE and S-TES-AE is statistically indistinguishable. Surprisingly, the D-VAE model
performed worse than both tree echo state models on all datasets,
which is likely caused by the small amount of training data, the short training time,
and, most of all, the lack of grammatical knowledge encoded in the network.
In particular, we observe that only $34\%$ of the decoded Boolean formulae,
$9\%$ of the decoded mathematical expressions, and none of the decoded
python programs conformed to the respective grammar. However, the architectural
bias of D-VAE was sufficient to at least achieve a tree structure for 
$100\%$ of the Boolean formulae, $95\%$ for the
mathematical expressions, and three of five python programs.

To check how training time influenced the results, we trained the D-VAE model
on the Boolean dataset again with $300$ epochs (just above $2.5$ hours of training time),
resulting in an RMSE of $3.70$ and $42\%$ grammatical correctness, which is still
considerably worse than the TES-AE model.

\begin{table}
\caption{Accuracy of all models in terms of RMSE on autoencoding the test data
($\pm$ standard deviation, except for D-VAE, which was evaluated only once)}
\label{tab:errors}
\begin{center}
\begin{tabular}{lcccc}
dataset & D-VAE & ES-AE & S-TES-AE & TES-AE \\
\cmidrule(lr){1-1} \cmidrule(lr){2-5}
Boolean & $4.62$ & $3.64 \pm 0.44$ & $3.25 \pm 0.39$ & $2.84 \pm 0.49$ \\
expressions & $5.81$ & $3.87 \pm 0.61$ & $2.65 \pm 0.23$ & $1.69 \pm 0.21$ \\
pysort & $52.07$ & $64.86 \pm 7.00$ & $16.97 \pm 4.30$ & $17.49 \pm 5.04$ \\
\end{tabular}
\end{center}
\end{table}

Regarding runtimes (refer to Table~\ref{tab:runtimes}), we observe that ES-AE
and S-TES-AE are comparably fast on the Boolean and pysort datasets but the
latter is factor 3 slower on the expressions dataset. Furthermore, TES-AE is
considerably slower on all datasets than S-TES-AE (factors ~1.5 on the first two
datasets and factor ~15 on the pysort dataset). This is to be expected as
setting up more parameters including a matrix decomposition for the spectral
radius computation for each parameter matrix is expensive. Further, the
parameter matrices for cycle reservoir with jumps \cite{Rodan2012} are sparser
than our Gaussian random number initialization, making ES-AE and S-TES-AE even
faster. In all cases, however, the overall runtime remains within a few seconds
time. This is in stark contrast to the D-VAE model, which took over 10 minutes
to train on the Boolean dataset, over 20 minutes for the expressions dataset,
and over 3 hours for the pysort dataset.

\begin{table}
\caption{Training time in seconds ($\pm$ standard deviation, except for
D-VAE which it was evaluated only once).}
\label{tab:runtimes}
\begin{center}
\begin{tabular}{lcccc}
dataset & D-VAE & ES-AE & S-TES-AE & TES-AE \\
\cmidrule(lr){1-1} \cmidrule(lr){2-5}
Boolean & $757.1$ & $1.26 \pm 0.04$ & $2.44 \pm 0.07$ & $3.29 \pm 0.22$ \\
expressions & $1201.76$ & $0.85 \pm 0.01$ & $3.37 \pm 0.10$ & $5.06 \pm 0.06$ \\
pysort & $13991.5$ & $0.47 \pm 0.02$ & $0.63 \pm 0.11$ & $10.83 \pm 0.76$ \\
\end{tabular}
\end{center}
\end{table}

\subsection{Optimization}

Next, we evaluated the capacity for tree optimization in the coding space.
For the Boolean dataset, we considered the logical evaluation of the formula,
assuming that $x$ is true and $y$ is false. We assign a score of $0$ if the formula evaluates to false and otherwise as the number of fulfilled
$\wedge$ terms in the formula. For example,
$\wedge(x, \neg(y))$ would evaluate to 1 because there is one fulfilled 'and'
but $\wedge(y, \wedge(x, x))$ would evaluate to 0 because the entire formula evaluates
to false.

For the expressions dataset, we used the performance measure of \cite{Kusner2017},
i.e.\ we evaluated the arithmetic expressions for $1000$
linearly spaced values of $x$ between $-10$ and $+10$ and computed the logarithm
of one plus the mean square error to the ground truth function
$1/3+x+\sin(x \cdot x)$.

Because our coding space was quite high-dimensional ($n = 256$), we did not
perform Bayesian optimization as suggested by \cite{Kusner2017} but used a
Covariance Matrix Adaptation Evolutionary Strategy (CMA-ES) instead, namely
the reference implementation of the python \texttt{cma} package\footnote{\url{https://github.com/CMA-ES/pycma}}.
To be comparable with \cite{Kusner2017}, we limited the computational
budget to the same value, namely $750$ overall function evaluations, which
we distributed onto 15 iterations with $50$ evaluations each.

The results are shown in Table~\ref{tab:optimization}. Note that the
results for D-VAE are missing because CMA-ES failed to generate any grammatical
tree which could have been evaluated. Regarding the results of the echo
state models, we note that the sequential echo state autoencoder (ES-AE)
performed best on the Boolean dataset by extrapolating beyond the training
data and using seven binary operators instead of the three that were present
in the training data. The TES-AE model also extrapolated, but with less success.
Only the S-TES-AE model remained within the boundaries of the training data and
achieved the best possible value within it.

Regarding the expression dataset, both TES-AE variations found a solution
at least as good as the grammar variational autoencoder of \cite{Kusner2017}
and the s-TES-AE model even found the ground truth. Overall, the s-TES-AE
model appears to be best suited for optimization on these tasks.

\begin{table}
\caption{The optimized tree and its score for all models for the Boolean
and expressions datasets. For Boolean, higher scores are better
and for expressions, lower scores are better.}
\label{tab:optimization}
\begin{center}
\begin{tabular}{lcc}
model & optimal expression & score \\
\cmidrule(lr){1-1} \cmidrule(lr){2-2} \cmidrule(lr){3-3}
& Boolean & \\
\cmidrule(lr){1-1} \cmidrule(lr){2-2} \cmidrule(lr){3-3}
ES-AE & $\wedge(\wedge(\vee(\wedge(x, x), x), x), \wedge(x, \wedge(x, \wedge(x, x))))$ & 6 \\
s-TES-AE & $\wedge(\wedge(\vee(y, x), x), \wedge(x, x))$ & 3 \\
TES-AE & $\vee(\neg(\vee(y, \wedge(\wedge(x, x), x))), \wedge(\wedge(\wedge(y, x)), x))$ & 3 \\
\cmidrule(lr){1-1} \cmidrule(lr){2-2} \cmidrule(lr){3-3}
& expressions & \\
\cmidrule(lr){1-1} \cmidrule(lr){2-2} \cmidrule(lr){3-3}
ES-AE & +(x, /(1, *(1, 3))) & 0.391 \\
s-TES-AE & +(/(1, 3), +(x, sin(*(x, x)))) & 0 \\
TES-AE & +(x, +(sin(3), sin(*(x, x)))) & 0.036 \\
\end{tabular}
\end{center}
\end{table}

\subsection{Coding Spaces}

\begin{figure}
\begin{center}
\begin{tikzpicture}
\begin{groupplot}[xticklabels={,,},yticklabels={,,},
group style={group size=2 by 1},
height=4cm,width=5cm
]
\nextgroupplot
\addplot[only marks, mark=*, fill=skyblue1, draw=skyblue3]
table[x=x,y=y,col sep=tab] {boolean_tesae_code_space.csv};
\node[left, outer sep=0.3cm] at (axis cs:-2.73439,-27.2817) {$\vee$};
\node[right, outer sep=0.3cm] at (axis cs:-23.0172,9.31803) {$\wedge$};
\node[left, outer sep=0.3cm] at (axis cs:14.2131,0.225303) {$y$};
\node[left, outer sep=0.3cm] at (axis cs:13.9007,28.689) {$x$};
\node[left, outer sep=0.3cm] at (axis cs:49.4786,-3.86115) {$\neg$};
\draw[semithick] (axis cs:40,5) rectangle (axis cs:57,-12);
\node[inner sep=0cm] (zoom_top) at (axis cs:40,5) {};
\node[inner sep=0cm] (zoom_bot) at (axis cs:40,-12) {};
\nextgroupplot
\addplot[only marks, mark=*, fill=skyblue1, draw=skyblue3]
table[x=x,y=y,col sep=tab] {boolean_tesae_code_space_not.csv};
\node[left, outer sep=0.3cm] at (axis cs:69.1167,-44.9931) {$\wedge$};
\node[right, outer sep=0.3cm] at (axis cs:-97.7742,-3.8061) {$\vee$};
\node[left, outer sep=0.3cm] at (axis cs:17.6364,31.6872) {$x$};
\node[below left, outer sep=0.2cm] at (axis cs:-10.9794,-27.4614) {$y$};
\end{groupplot}
\node (zoom_top2) at (4.5,+2.5) {};
\node (zoom_bot2) at (4.5,0) {};
\path[semithick]
(zoom_top) edge (zoom_top2)
(zoom_bot) edge (zoom_bot2);
\end{tikzpicture}
\end{center}
\caption{Left: A t-SNE visualization of the coding space of the TES-AE model on the
Boolean dataset. Each cluster in the visualization is labelled with the
root symbol of all trees in the cluster. Right: A t-SNE visualization of only
trees with $\neg$ at the root; clusters are labelled with the symbol below the root.}
\label{fig:coding_space}
\end{figure}
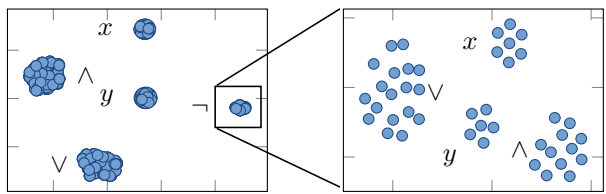

If we inspect the encoding spaces of the TES-AE model in more detail, we observe
clusters dependent on root symbol of the tree. For example,
Figure~\ref{fig:coding_space} (left) shows a t-SNE dimensionality reduction
of the Boolean dataset as encoded by the TES-AE model with each cluster labelled with
the root symbol. In Figure~\ref{fig:coding_space} (right), we observe that the $\neg$
cluster further spreads into clusters according to the symbol just below the root.
This fractal coding is consistent with prior work on recurrent networks with small weights,
which have been shown to code fractally based on the most recent symbol \cite{Tino2003}.

\section{Conclusion}

In this paper, we introduced tree echo state autoencoders (TES-AE), a novel neural network
architecture to implement autoencoding for trees without the need for deep learning.
In particular, we used regular tree grammars to express our trees as sequences of
grammar rules and then employed echo state networks and tree echo state networks for
encoding and decoding.
In our experiments on three datasets, we found that a TES-AE outperformed a variational
auto-encoder for acyclic graphs (D-VAE) in terms of autoencoding error on small datasets
with limited training time.
Further, we showed that TES-AE significantly outperform a sequential version
of the model (ES-AE) and that separate parameters for each grammar rule outperform
shared parameters. Our results also showed that a few seconds sufficed to train our model even
for a large grammar and large trees, whereas D-VAE training, even with a small number
of epochs, took ten minutes to several hours. Finally, we observed that optimization in the
TES-AE coding space performed similarly compared to past reference results \cite{Kusner2017}.

Future research could investigate how well our autoencoders are suitable to time series
prediction, how memory capacity results translate to the tree domain, how to apply our architecture
to trees with real-valued nodes, and whether our proposed echo state
sequence-to-sequence learning model using echo state networks is suitable to solve sequence tasks
that currently require deep learning.

\bibliography{literature} 
\bibliographystyle{plainnat}

\end{document}